%% file: main.tex
\documentclass[10pt,twocolumn,letterpaper]{article}

\usepackage[pagenumbers]{iccv} 

\input{preamble}

\definecolor{iccvblue}{rgb}{0.21,0.49,0.74}
\usepackage[pagebackref,breaklinks,colorlinks,allcolors=iccvblue]{hyperref}

\def\paperID{1085} 
\def\confName{ICCV}
\def\confYear{2025}

\title{DiT4SR: Taming Diffusion Transformer for Real-World Image Super-Resolution}

\author{Zheng-Peng Duan\textsuperscript{1,2}
\thanks{This project is done during the internship at SenseTime Research.}
\and
Jiawei Zhang\textsuperscript{2}
\and
Xin Jin\textsuperscript{1}
\and
Ziheng Zhang\textsuperscript{1}
\and
Zheng Xiong\textsuperscript{2}
\and
Dongqing Zou\textsuperscript{2,3}
\and
Jimmy S. Ren\textsuperscript{2,4}
\and
Chunle Guo\textsuperscript{1}
\and
Chongyi Li\textsuperscript{1}
\thanks{Corresponding author.}
\and
\small{{\textsuperscript{1}VCIP, CS, Nankai University}} \quad
\small{{\textsuperscript{2}SenseTime Research}} \quad
\small{{\textsuperscript{3}PBVR}} \quad
\small{{\textsuperscript{4}Hong Kong Metropolitan University}} \\
}

\begin{document}
\maketitle

\begin{strip}
\begin{minipage}{\textwidth}
\vspace{-16mm}
\centering
\includegraphics[width=0.98\linewidth]{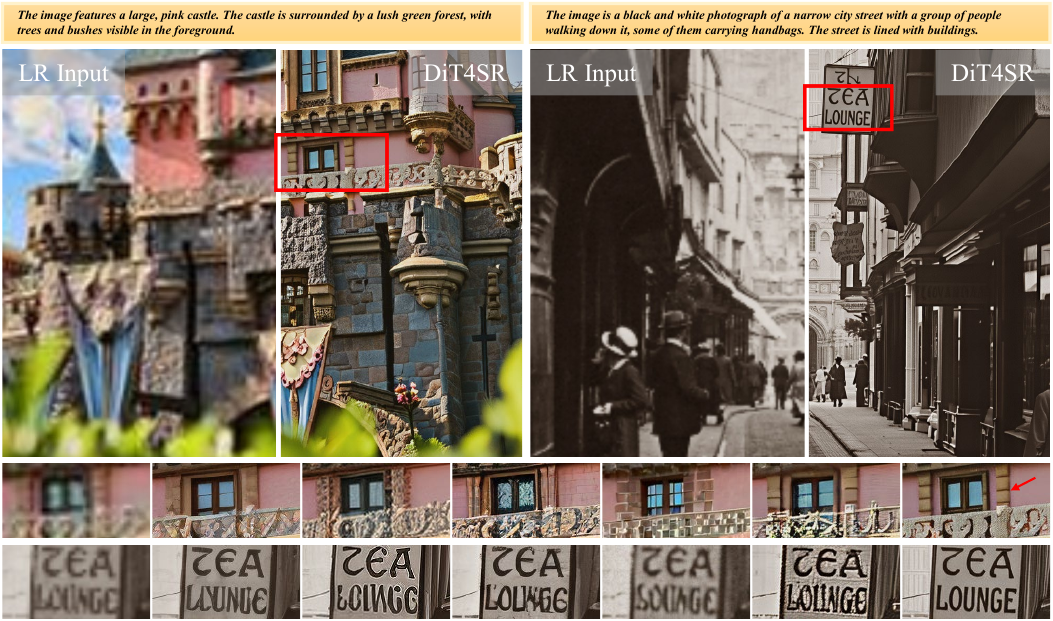}
\put(-468.0, -11.0){LR Input}
\put(-404.0, -11.0){SeeSR~\cite{wu2024seesr}}
\put(-338.0, -11.0){DiffBIR~\cite{lin2024diffbir}}
\put(-276.0, -11.0){DreamClear~\cite{ai2025dreamclear}}
\put(-198.0, -11.0){SUPIR~\cite{yu2024scaling}}
\put(-140.0, -11.0){SD3-CtrlNet~\cite{zhang2023adding}}
\put(-50.0, -11.0){DiT4SR}
\vspace{-2mm}
\captionof{figure}{Performance comparison between our DiT4SR and other state-of-the-art methods on two real-world LR images. Thanks to the powerful generative capabilities of SD3 and our delicate designs, which allow sufficient interaction between the input LR information and the generation process, our method produces more realistic details while maintaining fidelity, especially in text, structures, and details.}
\label{fig:teaser}
\vspace{-2mm}

\end{minipage}
\end{strip}

\input{sec/0_abstract}    
\input{sec/1_intro}
\input{sec/2_formatting}
{
    \small
    \bibliographystyle{ieeenat_fullname}
    \bibliography{main}
}

\end{document}

%% file: preamble.tex
\usepackage{amsmath}
\usepackage{amssymb}
\usepackage{mathtools}
\usepackage{amsthm}
\usepackage{bbding}

\usepackage{cuted}
\usepackage{caption}
\usepackage{subcaption}
\usepackage{multirow}
\usepackage{tikz}

\newcommand{\mathimg}[1]{\mathbf{#1}}

\newcommand{\figref}[1]{Figure~\ref{#1}}
\newcommand{\tabref}[1]{Table~\ref{#1}}
\newcommand{\secref}[1]{Section~\ref{#1}}

\newcommand*\circled[1]{\tikz[baseline=(char.base)]{
            \node[shape=circle,draw,inner sep=1pt] (char) {#1};}}

\newcommand{\myparaheadd}[1]{\noindent\textbf{#1}}

%% file: sec/0_abstract.tex
\begin{abstract}
Large-scale pre-trained diffusion models are becoming increasingly popular in solving the Real-World Image Super-Resolution (Real-ISR) problem because of their rich generative priors.
The recent development of diffusion transformer (DiT) has witnessed overwhelming performance over the traditional UNet-based architecture in image generation,
which also raises the question: 
Can we adopt the advanced DiT-based diffusion model for Real-ISR?
To this end,
we propose our DiT4SR, 
one of the pioneering works to tame the large-scale DiT model for Real-ISR.
Instead of directly injecting embeddings extracted from low-resolution (LR) images like ControlNet,
we integrate the LR embeddings into the original attention mechanism of DiT, 
allowing for the bidirectional flow of information between the LR latent and the generated latent.
The sufficient interaction of these two streams allows the LR stream to evolve with the diffusion process, producing progressively refined guidance that better aligns with the generated latent at each diffusion step.
Additionally, the LR guidance is injected into the generated latent via a cross-stream convolution layer, compensating for DiT's limited ability to capture local information.
These simple but effective designs endow the DiT model with superior performance in Real-ISR,
which is demonstrated by extensive experiments.
Project Page: \href{https://adam-duan.github.io/projects/dit4sr/}{https://adam-duan.github.io/projects/dit4sr/}.
\end{abstract}

%% file: sec/1_intro.tex
\section{Introduction}
\label{sec:intro}
Real-ISR~\cite{zhang2021designing, wang2021real} aims to recover a high-resolution (HR) image from its LR version under various types of degradations,
such as compression, blur, and noise.
Unlike traditional ISR~\cite{dong2014learning},
Real-ISR requires that the model can not only remove the complex degradations,
but also generate perceptually realistic details to enhance the visual quality.
The high ill-posedness of this task places demands on prior knowledge of the model.
Without such prior,
the model is prone to generating ambiguous results and suffers from degraded restoration quality.
Therefore, researchers have turned their attention to large-scale pre-trained text-to-image (T2I) models~\cite{saharia2022photorealistic, chen2023pixart}, 
especially Stable Diffusion (SD), which are trained on billions of high-quality nature images and encompass rich real-world prior knowledge.
They build their methods upon UNet-based SD, known as SD1~\cite{rombach2022high}, SD2~\cite{rombach2022high}, and SDXL~\cite{podell2023sdxl},
and treat input LR images as conditions via ControlNet~\cite{zhang2023adding} or ControlNet-like manners to generate the corresponding HR images.

Recently, 
the development of diffusion transformer (DiT)~\cite{peebles2023scalable} has made DiT-based architecture prevalent.
%
Owing to the overwhelming performance of SD3~\cite{esser2024scaling} and Flux~\cite{blackforestlabs2024}, 
MM-DiT has been proven to be an effective transformer block for generative models.
It features two learnable streams for visual features and text tokens and enables a bidirectional flow of information between the two modalities through attention operation.
Building on such architectural innovations, along with other techniques like scaling up,
DiT-based diffusion models demonstrate promising performance in detail generation and image quality.
\textit{Given these merits,
a pivotal question arises:
Can we adopt the advanced DiT-based diffusion model for Real-ISR?}

\begin{figure}[t]
\begin{center}
\includegraphics[width=.98\columnwidth]{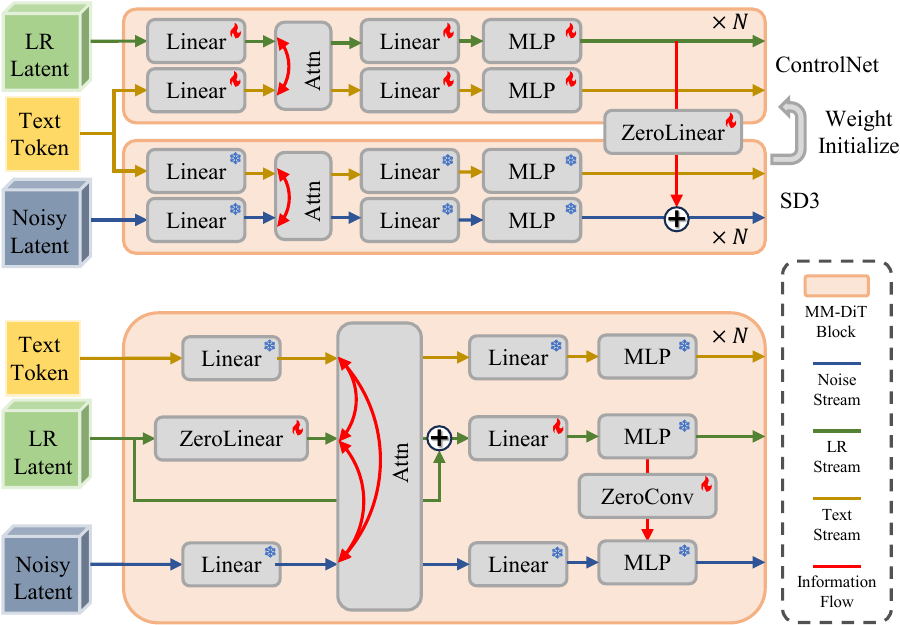}
\put(-162.0, 86.0){\footnotesize{(a) SD3-ControlNet~\cite{zhang2023adding}}}
\put(-142.0, -10.0){\footnotesize{(b) DiT4SR}}
\vspace{-2mm}
\caption{Network Structure comparison between SD3-ControlNet and our DiT4SR. 
The information flow across streams is marked with \textcolor{red}{red} lines and the direction is indicated by arrows. Notably, our DiT4SR enables bidirectional information interaction, allowing the LR Stream and Noise Stream to continuously interact and evolve together, whereas SD3-ControlNet relies on one-direction information flow, restricting the interaction. }
\label{fig:teaser2}
\end{center}
\vspace{-10mm}
\end{figure}

An intuitive solution is to adopt ControlNet~\cite{zhang2023adding} as the DiT controller
and inject LR information to guide the generative process for Real-ISR,
which is shown in \figref{fig:teaser2}~(a).\footnote{The implementation of SD3-ControlNet follows the default code and configuration provided by Diffusers~\cite{von-platen-etal-2022-diffusers}.} 
Specifically,
several MM-DiT blocks are duplicated as a trainable copy to initialize ControlNet.
%
%
The LR latent, obtained by passing the LR image through the pre-trained VAE encoder, is used as input. 
The hidden state output from each block in ControlNet is then directly added to the Noise Stream of SD3 through a trainable convolution layer.
%
%
%
%
Since ControlNet is originally designed for UNet-based architectures, directly adding LR information ignores the unique characteristics of DiT, restricting information interactions and potentially limiting performance.

However, it is non-trivial to tame the DiT-based diffusion model for Real-ISR. 
To leverage the strength of DiT, we abandon the traditional technical roadmap such as ControlNet and propose a new control architecture designed specifically for DiT, named DiT4SR.
Instead of processing the LR information in the additionally created DiT blocks like~\figref{fig:teaser2}~(a),
we integrate the LR Stream into the DiT blocks, allowing for more efficient interaction between LR information and the diffusion process.
%
%
%
%
Considering that both the LR latent and the noisy latent are forms of visual information,
we adopt a similar design for the LR Stream as the Noise Stream.
%
%
In each MM-DiT block,
our method duplicates the modules for the Noise Stream to process the LR input.
%
%
Such \textbf{LR Integration in Attention} allows for the bidirectional information interaction between the LR Stream and the Noise Stream,
which is shown in \figref{fig:teaser2}~(b).

The bidirectional interaction between the two streams allows the LR Stream to evolve alongside the diffusion process, enabling it to generate progressively refined and context-aware guidance that better aligns with the generated latent.
%
%
%
To preserve LR information and enhance consistency in deeper blocks, 
an additional \textbf{LR Residual} is introduced to directly connect the input and output of the attention module within the LR Stream.
%
%
However, since the attention mechanism operates globally, it alone is not sufficient for SR tasks. 
Capturing local information is equally essential to restore fine details.
%
%
Therefore,
we further inject the LR guidance into the Noise Stream through the convolution layer.
The \textbf{LR Injection between MLP} allows our model to aggregate the local information, compensating for the limited local information-capturing ability of DiT~\cite{xie2024sana}.
%


Our \textbf{contributions} can be summarized as follows:
\begin{itemize}
    \item To our knowledge, our method is one of the pioneering works to tame the large-scale DiT model for Real-ISR.
    \item Instead of copying blocks like ControlNet, we integrate the LR Stream into the original DiT block,
    which enables bidirectional information interaction between the LR guidance and the diffusion process.
    \item We introduce a convolutional layer for injecting LR guidance into the Noise Stream, which compensates for the limited local information-capturing ability of DiT.
\end{itemize}

%% file: sec/2_formatting.tex
\begin{figure*}[t]
\begin{center}
\centerline{\includegraphics[width=.96\linewidth]{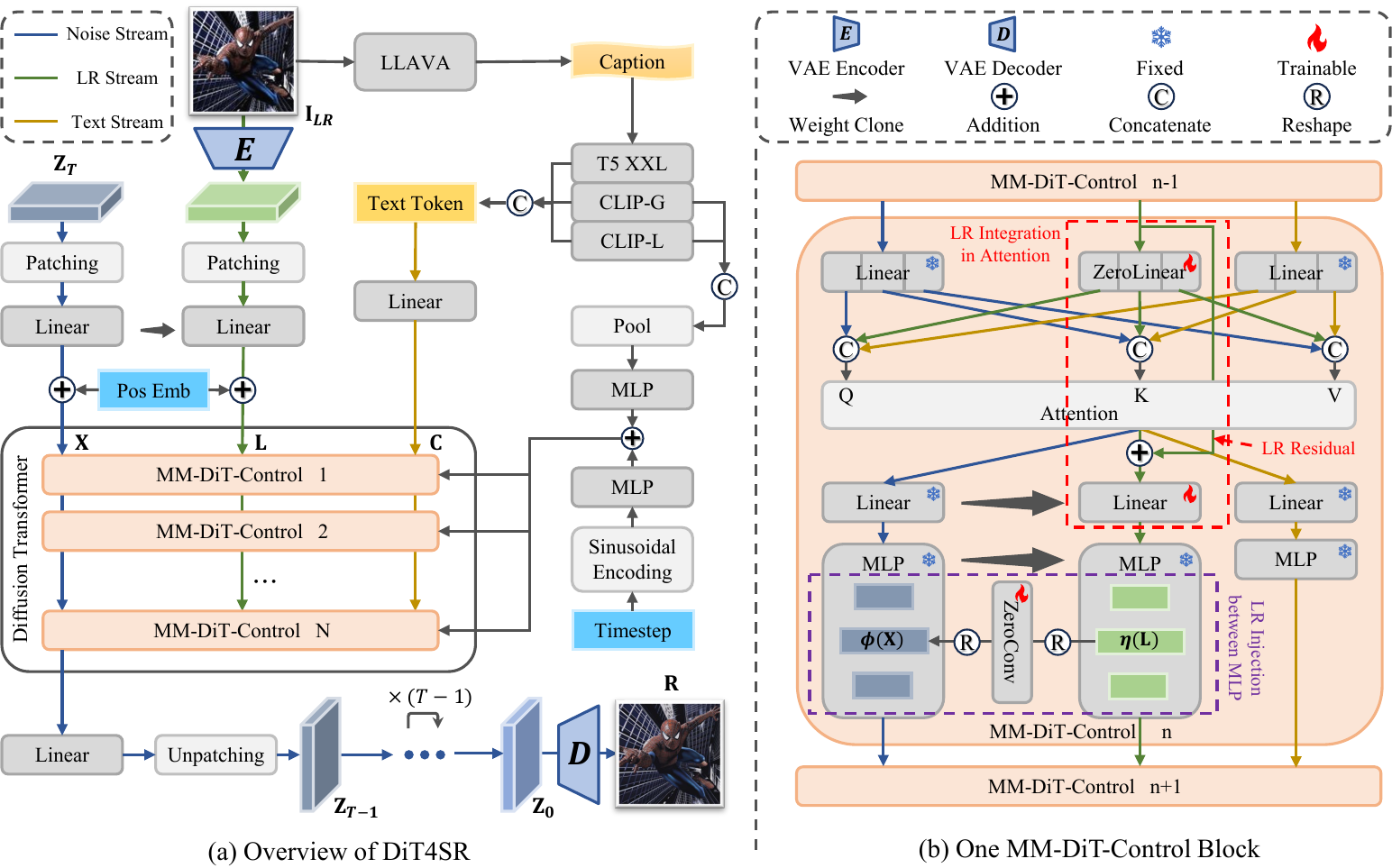}}
\caption{Overview of DiT4SR (a) and MM-DiT-Control block (b). Our DiT4SR integrates LR Stream to MM-DiT-Control Block. \textbf{LR Integration in Attention} allows the bidirectional interaction between LR Stream and Noise Stream. 
\textbf{LR Residual} is introduced to enhance the consistency of LR guidance.
Additionally, \textbf{LR Injection between MLP} injects the LR guidance from LR Stream to Noise Stream through a
trainable convolution layer.}
\label{fig:pipeline}
\end{center}
\vspace{-11mm}
\end{figure*}

\section{Related Work}
\myparaheadd{Image Super-Resolution}
%
Deep learning-based ISR methods have achieved significant progress,
where the architectures develop from convolutional networks~\cite{dong2014learning, dai2019second, lim2017enhanced, zhang2022efficient, zhang2018image, zhang2018residual} to transformers~\cite{chen2021pre, chen2023activating, chen2023dual, liang2021swinir}.
However,
these methods struggle to handle Real-ISR due to the complex degradations in real-world scenarios and the ill-posed nature of this task.
To address the former issue,
BSRGAN~\cite{zhang2021designing} and Real-ESRGAN~\cite{wang2021real} explore more complex degradation models.
For the ill-posed property,
several GAN-based methods~\cite{liang2022efficient, wang2018esrgan, chen2023human, chan2021glean} are proposed to learn the distribution of high-quality images and generate perceptually realistic details.
However, these methods still suffer from training instability and generate unnatural visual artifacts~\cite{chen2022real, liang2022details, xie2023desra}.

With the success of large-scale T2I models such as Stable Diffusion (SD)~\cite{rombach2022high},
researchers~\cite{qu2024xpsr, yue2024resshift, wan2024clearsr, gu2024consissr, sun2024coser, cui2024taming, tsao2024holisdip, sun2024pixel, li2024foundir} have shifted their attention to use the generative priors embedded in these powerful pre-trained models to handle the Real-ISR problem.
%
%
Built upon the first-generation SD~\cite{rombach2022high},
which is based on a UNet architecture, 
StableSR~\cite{wang2024exploiting} and DiffBIR~\cite{lin2024diffbir} inject the LR information as conditions via ControlNet~\cite{zhang2023adding} or ControlNet-like manners.
%
PASD~\cite{yang2024pixel} and SeeSR~\cite{wu2024seesr} further incorporate high-level semantic information to guide the diffusion process.
%
Employing the more advanced SDXL~\cite{podell2023sdxl},
SUPIR investigates the effects of scaling up in Real-ISR.
There are also some other methods~\cite{wang2024sinsr, zhang2024degradation, li2024distillation, xie2024addsr, chen2024adversarial, wu2024one,sun2024improving, liu2024patchscaler, noroozi2024you} committed to a more efficient diffusion process.
With the development of diffusion transformer (DiT),
DiT-SR~\cite{cheng2025ditsr} trains DiT-based SR models from scratch.
%
DreamClear~\cite{ai2025dreamclear} proposes a DiT-based image restoration model, 
but still adopts the ControlNet to inject LR information,
which prevents it from fully leveraging the advantage of DiT.
%
From the perspective of network architecture, our DiT4SR is one of the pioneering works to tame the large-scale diffusion transformer model for Real-ISR,
while discarding the ControlNet-like approach.
There are two concurrent methods~\cite{li2025fluxsr, dong2025tsd} that explore one-step DiT-based SR models in terms of flow trajectory.

\myparaheadd{Diffusion Transformer}
To enhance the generative capability of diffusion models,
large-scale transformer architectures have been introduced,
where diffusion transformer (DiT)~\cite{peebles2023scalable} stands out.
Building on DiT,
large-scale T2I models,
\eg PixArt-$\alpha$ ~\cite{chen2023pixart}, SD3~\cite{esser2024scaling}, and Flux~\cite{blackforestlabs2024},
are proposed.
Specifically,
SD3 and Flux leverage Multimodal Diffusion Transformers (MM-DiTs) to integrate text and image modalities through attention operation.
In this way, the two modalities can fully interact, forming the core advantage of DiT.
Our DiT4SR further enhances this advantage by incorporating the LR stream into the DiT blocks, enabling sufficient interaction between LR information and original features within the DiT blocks.
%
%

\section{Methodology}
Given a LR image $\mathimg{I}_{LR}$ undergoing complex degradations,
Real-ISR aims to recover the corresponding HR image $\mathimg{R}$.
Along with the removal of degradations, 
Real-ISR simultaneously requires the model to generate realistic details,
thereby enhancing visual perception.
To overcome the ill-posedness of this task,
researchers build their methods on large-scale pre-trained T2I models, especially Stable Diffusion (SD), to leverage the rich generative priors.
The recent development of diffusion transformer (DiT) has gradually replaced the traditional UNet-based diffusion model,
achieving state-of-the-art performance in detail generation and visual quality.
In this study,
we focus on leveraging the strength of DiT and tame it for Real-ISR.

To this end,
we propose our DiT4SR,
which is built on DiT-based SD3.
The overview of the architecture of our DiT4SR is introduced in \secref{sec:arc}.
Instead of creating additional DiT blocks to process the LR information like SD3-ControlNet,
we integrate the LR Stream into the original attention calculation of DiT and allow bidirectional interaction of information with the Noise Stream and Text Stream,
which is detailed in \secref{sec:attn}.
However, relying solely on the attention mechanism is not enough to make the DiT competent for Real-ISR.
To inject more LR guidance and capture the local information of LR features,
the intermediate features of MLP in the LR Stream are introduced into that of the Noise Stream through a depth-wise convolution layer, 
which are detailed in \secref{sec:mlp}.

\subsection{Overview of Architecture}
\label{sec:arc}
Since our DiT4SR is built on the popular DiT-architectured SD3,
we first introduce SD3 briefly.
Like previous SD models~\cite{rombach2022high},
SD3 also conducts the diffusion process in the latent space.
%
It consists of a sequence of MM-DiT blocks, where separate weight sets handle text and image embeddings, forming the Noise Stream and the Text Stream,
as shown in the bottom part of \figref{fig:teaser2}~(a).
%
This structure employs two independent transformers for each modality while merging their sequences during the attention mechanism, enabling cross-modal interaction.
The bidirectional interaction enables both streams to evolve together throughout the diffusion process, representing the key strength of the DiT.
If ControlNet \cite{zhang2023adding, von-platen-etal-2022-diffusers} is adopted as the DiT controller for Real-ISR, as illustrated in \figref{fig:teaser2} (a), SD3-ControlNet processes the LR Stream within additional DiT blocks and directly injects the LR embedding into the Noise Stream via trainable linear layers.
%
This approach establishes only a one-way information flow from the LR Stream to the Noise Stream, limiting information interaction.
In contrast, our DiT4SR integrates the LR Stream directly into the original DiT blocks, as illustrated in \figref{fig:teaser2} (b). 
%
This design enables bidirectional information flow, allowing the LR Stream to continuously adapt throughout the diffusion process and generate guidance that aligns more effectively with the evolving Noise Stream.

As illustrated in \figref{fig:pipeline},
our DiT4SR shares a similar architecture with SD3.
Specifically, before being fed into the diffusion transformer, the noisy latent $\mathimg{Z}\in \mathbb{R}^{H\times W\times C}$ is first flattened into a patch sequence of length $K$, where $K = \frac{H}{2} \cdot \frac{W}{2}$, and then projected to a $D$-dimensional space using a linear layer.
Then, the position embedding is added to the obtained noisy image token $\mathimg{X} \in \mathbb{R}^{K \times D}$.
To introduce the LR information,
we first encode the LR image $\mathimg{I}_{LR}$ into the latent space through the pre-trained VAE encoder.
Considering the LR latent and noisy latent are both forms of visual representations, 
we follow the same process and add the same position embedding to obtain the LR image token $\mathimg{L}\in \mathbb{R}^{K \times D}$.
Following SD3~\cite{esser2024scaling},
the input caption that describes $\mathimg{I}_{LR}$ is encoded by three pre-trained text models, including CLIP-L~\cite{radford2021learning}, CLIP-G~\cite{cherti2023reproducible}, and T5 XXL~\cite{raffel2020exploring}.
The text representations output by two CLIP models are pooled and combined with the timestep $t$ to modulate the internal features of DiT.
Besides the pooled text representations,
the text token $\mathimg{C}\in \mathbb{R}^{M \times D}$ is also constructed with a length of $M$ by combining all three text representations.

With the noisy image token $\mathimg{X}$ and the text token $\mathimg{C}$ as input,
the original MM-DiT block adopts two separate sets of weights for these two modalities, named Noise Stream and Text Stream.
In our DiT4SR,
we introduce an additional stream, named LR Stream, to tackle the LR image token $\mathimg{L}$.
The MM-DiT block is modified as the MM-DiT-Control block,
which allows the LR information to guide the generation of HR latent.
After $N$ MM-DiT-Control blocks and the unpatch operation,
Noise Stream outputs the denoised latent for this timestep $t$.
Repeating the diffusion process for $T$ steps and decoding the clean latent $\mathimg{Z}_0$,
we can finally obtain the desired HR result $\mathimg{R}$.
In the following two sections,
the detailed modifications in the MM-DiT-Control block are introduced.

\subsection{LR Integration in Attention}
\label{sec:attn}
\begin{figure}[t]
\begin{center}
\centerline{\includegraphics[width=.98\columnwidth]{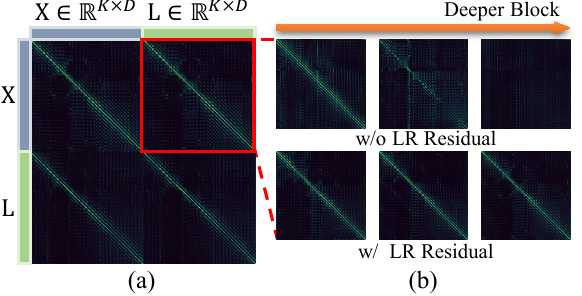}}
\caption{(a) Visualization of four attention maps for noisy image token $\mathimg{X}$ and LR image token $\mathimg{L}$ ($\mathimg{X} \rightarrow\mathimg{X}$, $\mathimg{X} \rightarrow\mathimg{L}$, $\mathimg{L} \rightarrow\mathimg{X}$, $\mathimg{L} \rightarrow\mathimg{L}$) in the 24th MM-DiT-Control. Diagonal lines of $\mathimg{X} \rightarrow\mathimg{L}$ and $\mathimg{L} \rightarrow\mathimg{X}$ indicate the information interaction between $\mathimg{X}$ and $\mathimg{L}$. (b) Attention maps for $\mathimg{X} \rightarrow\mathimg{L} $ w/ and w/o LR Residual. Without LR Residual, LR guidance diminishes with increasing block depth (in the 1st, 13th, and 24th MM-DiT-Control). LR Residual can significantly enhance the consistency of LR guidance. }
\label{fig:attn}
\end{center}
\vspace{-11mm}
\end{figure}

Following the design of the original MM-DiT,
all three streams sequentially undergo the joint attention mechanism and the MLP operation in the  MM-DiT-Control block.
As for the joint attention,
the input can be formulated as
\begin{equation}
\begin{aligned}
    \mathimg{Q} = P_\mathimg{Q}^\mathimg{X}(\mathimg{X})~\circled{c}~P_\mathimg{Q}^\mathimg{L}(\mathimg{L})~\circled{c}~P_\mathimg{Q}^\mathimg{C}(\mathimg{C}), \\ 
    \mathimg{K} = P_\mathimg{K}^\mathimg{X}(\mathimg{X})~\circled{c}~P_\mathimg{K}^\mathimg{L}(\mathimg{L})~\circled{c}~P_\mathimg{K}^\mathimg{C}(\mathimg{C}), \\ 
    \mathimg{V} = P_\mathimg{V}^\mathimg{X}(\mathimg{X})~\circled{c}~P_\mathimg{V}^\mathimg{L}(\mathimg{L})~\circled{c}~P_\mathimg{V}^\mathimg{C}(\mathimg{C}), \\ 
\end{aligned}
\end{equation}
where $P_\mathimg{Q}^\mathimg{X}$, $P_\mathimg{K}^\mathimg{X}$, $P_\mathimg{V}^\mathimg{X}$, $P_\mathimg{Q}^\mathimg{C}$, $P_\mathimg{K}^\mathimg{C}$, and $P_\mathimg{V}^\mathimg{C}$ are pre-trained fixed linear projections for $\mathimg{X}$ and $\mathimg{C}$.
$\circled{c}$ denotes the operation of concatenation in the token length dimension.
$P_\mathimg{Q}^\mathimg{L}$, $P_\mathimg{K}^\mathimg{L}$, and $P_\mathimg{V}^\mathimg{L}$ are newly created trainable linear projections for $\mathimg{L}$,
whose weights are initialized to zeros.
In this way, 
the impact of $\mathimg{L}$ can be ignored at the beginning of the training, and progressively grow during the training.
The joint attention in MM-DiT-Control is calculated as
\begin{equation}
\text{Attention}(\mathimg{Q}, \mathimg{K}, \mathimg{V}) = \underbrace{\text{softmax}(\frac{\mathimg{Q}\mathimg{K}^T}{\sqrt{d}})}_{\text{attention map}}\mathimg{V},
\end{equation}
which allows for more comprehensive interaction between these three streams.
We further visualize the attention maps regarding LR image token $\mathimg{X}$ and noisy image token $\mathimg{L}$ in \figref{fig:attn} (a).
We observe that the diagonals are activated obviously in both self-attention regions and cross-attention regions,
which indicates the information interaction between the corresponding positions on $\mathimg{X}$ and $\mathimg{L}$.
This bidirectional interaction not only enables the Noise Stream to be influenced by LR guidance but also allows the LR Stream to adapt based on the state of the Noise Stream, providing more accurate and context-aware guidance.

Another noteworthy thing is the information interaction between $\mathimg{L}$ and $\mathimg{X}$ decay through successive attention blocks,
which is shown in \figref{fig:attn} (b).
%
%
The possible reason is that, under the influence of joint attention, the LR information evolves alongside the Noise Stream and may experience undesired disruptions. 
This gradual degradation reduces its effectiveness in guiding the Noise Stream.
To enhance the consistency of LR guidance, an additional shortcut directly transfers the input LR information to the output of the joint attention mechanism. 
By incorporating the \textbf{LR Residual}, the LR guidance is effectively preserved in deeper transformer blocks, ensuring its consistent influence on the Noise Stream throughout the diffusion process.
The visual comparison between the models with and without LR Residual is presented in~\figref{fig:ablation} (c) and (f).

\subsection{LR Injection between MLP}
\label{sec:mlp}
\begin{figure}[t]
\begin{center}
\centerline{\includegraphics[width=.96\columnwidth]{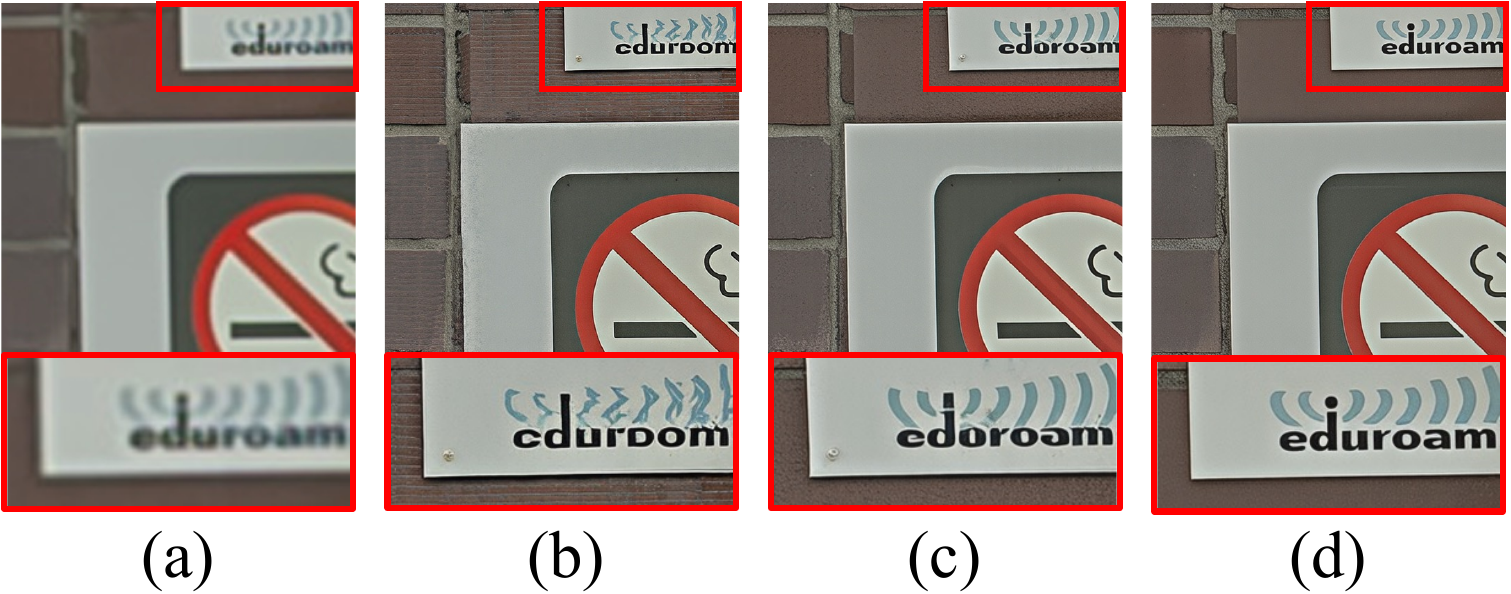}}
\caption{(a) is the LR input. (b) is the result w/o LR Injection between MLP. (c) injects the LR information through a linear layer. (d) is the result of our DiT4SR which replaces the linear layer with a convolution layer. The convolution layer helps capture more precise local information, reflected by remarkable performance in recovering fine structures. All results are obtained by retraining the specific models under the same settings.}
\label{fig:ffplus}
\end{center}
\vspace{-13mm}
\end{figure}

Since joint attention operates at a global level and relies solely on position embeddings to provide spatial information, it is insufficient for effectively adapting DiT to Real-ISR. 
%
%
This limitation arises because local information is equally important for restoring fine details, which global attention alone may overlook.
As shown in \figref{fig:ffplus} (b),
relying solely on the joint attention still faces a challenge in restoring the text and maintaining image fidelity.

To strengthen the guidance of LR information,
we further connect the intermediate features of the MLPs from LR Stream to Noisy Stream with a $3\times 3$ depth-wise convolution layer whose weights are initialized to zeros.
Specifically, in both MLPs for the LR Stream and Noisy Stream, the hidden state dimensions are first expanded by a factor of 4 and then projected back to the original size through two linear projections.
We denote these two intermediate features as $\phi(\mathimg{X})\in \mathbb{R}^{K \times 4D}$ and $\eta(\mathimg{L})\in \mathbb{R}^{K \times 4D}$.
We first reshape $\eta(\mathimg{L})$ to $\eta(\mathimg{L})'\in  \mathbb{R}^{\frac{H}{2} \times \frac{W}{2} \times 4D}$,
and then pass it through the $3\times 3$ depth-wise convolution layer.
After reshaping back to the image token form, the LR information is effectively injected into the Noise Stream.

\begin{table*}[t]
\footnotesize
\renewcommand{\arraystretch}{1.1}
\setlength{\tabcolsep}{4.0pt}
\centering
\begin{tabular}{c|c|cc|cccccc|ccc}
\toprule
Datasets                    & Metrics & \begin{tabular}[c]{@{}c@{}}Real-\\ ESRGAN\end{tabular} & SwinIR                       & ResShift & StableSR                     & SeeSR                         & DiffBIR          & OSEDiff             & SUPIR  & DreamClear & \begin{tabular}[c]{@{}c@{}}SD3-\\ ControlNet\end{tabular} & DiT4SR                        \\ 
\midrule
                            & LPIPS $\downarrow$   & 0.282                                                  & {\color[HTML]{2972F4} 0.274} & 0.353    & {\color[HTML]{FF0000} 0.273} & 0.317                         & 0.452                  &   0.297    & 0.419  & 0.354      & 0.323                                                     & 0.365                         \\
                            & MUSIQ $\uparrow$   & 54.267                                                 & 52.737                       & 52.392   & 58.512                       & {\color[HTML]{2972F4} 65.077} & {\color[HTML]{FF0000} 65.665} & 64.692 & 59.744 & 44.047     & 55.956                                                    & 64.950                         \\
                            & MANIQA $\uparrow$  & 0.490                                                   & 0.475                        & 0.476    & 0.559                        & 0.605                         & {\color[HTML]{FF0000} 0.629} & 0.590 & 0.552  & 0.455      & 0.545                                                     & {\color[HTML]{2972F4} 0.627}  \\
                            & ClipIQA $\uparrow$ & 0.409                                                  & 0.396                        & 0.379    & 0.438                        & 0.543                         & {\color[HTML]{FF0000} 0.572} & 0.519 & 0.518  & 0.379      & 0.449                                                     & {\color[HTML]{2972F4} 0.548}  \\
\multirow{-5}{*}{DrealSR}   & LIQE $\uparrow$    & 2.927                                                  & 2.745                        & 2.798    & 3.243                        & {\color[HTML]{FF0000} 4.126}  & 3.894                       & 3.942 & 3.728  & 2.401      & 3.059                                                     & {\color[HTML]{3166FF} 3.964}  \\
\midrule
                            & LPIPS $\downarrow$   & {\color[HTML]{2972F4} 0.271}                           & {\color[HTML]{FF0000} 0.254} & 0.316    & 0.306                        & 0.299                         & 0.347                     &  0.292  & 0.357  & 0.325      & 0.305                                                     & 0.319                         \\
                            & MUSIQ $\uparrow$   & 60.370                                                  & 58.694                       & 56.892   & 65.653                       & {\color[HTML]{FF0000} 69.675} & 68.340 & {\color[HTML]{2972F4} 69.087 } & 61.929 & 59.396     & 62.604                                                    & 68.073                        \\
                            & MANIQA $\uparrow$  & 0.551                                                  & 0.524                        & 0.511    & 0.622                        & 0.643                         & {\color[HTML]{2972F4} 0.653} & 0.634 & 0.574  & 0.546      & 0.599                                                     & {\color[HTML]{FF0000} 0.661}  \\
                            & ClipIQA $\uparrow$ & 0.432                                                  & 0.422                        & 0.407    & 0.472                        & {\color[HTML]{2972F4} 0.577}  & {\color[HTML]{FF0000} 0.586} & 0.552 & 0.543  & 0.474      & 0.484                                                     & 0.550                          \\
\multirow{-5}{*}{RealSR}    & LIQE $\uparrow$    & 3.358                                                  & 2.956                        & 2.853    & 3.750                         & {\color[HTML]{FF0000} 4.123}  & 4.026 & {\color[HTML]{2972F4} 4.065} & 3.780   & 3.221      & 3.338                                                     & 3.977                         \\
\midrule
                            & MUSIQ $\uparrow$   & 62.961                                                 & 63.548                       & 59.695   & 63.433                       & 69.428 & 68.027                      & {\color[HTML]{2972F4} 69.547}  & 64.837 & 65.926     & 65.623                                                    & {\color[HTML]{FF0000} 70.469} \\
                            & MANIQA $\uparrow$  & 0.553                                                  & 0.560                         & 0.525    & 0.579                        & 0.612                         & {\color[HTML]{2972F4} 0.629} & 0.606 & 0.600    & 0.597      & 0.587                                                     & {\color[HTML]{FF0000} 0.645}  \\
                            & ClipIQA $\uparrow$ & 0.451                                                  & 0.463                        & 0.452    & 0.458                        & 0.566                         & {\color[HTML]{2972F4} 0.582} & 0.551 & 0.524  & 0.546      & 0.526                                                     & {\color[HTML]{FF0000} 0.588}  \\
\multirow{-4}{*}{RealLR200} & LIQE $\uparrow$    & 3.484                                                  & 3.465                        & 3.054    & 3.379                        & 4.006  & 4.003                        & {\color[HTML]{2972F4} 4.069} & 3.626  & 3.775      & 3.733                                                     & {\color[HTML]{FF0000} 4.331}  \\
\midrule
                            & MUSIQ $\uparrow$   & 62.514                                                 & 63.371                       & 59.337   & 56.858                       & {\color[HTML]{2972F4} 70.556} & 69.876                       & 69.580 & 66.016 & 66.693     & 66.385                                                    & {\color[HTML]{FF0000} 71.832} \\
                            & MANIQA $\uparrow$  & 0.524                                                  & 0.534                        & 0.500      & 0.504                        & 0.594                         & {\color[HTML]{2972F4} 0.624} & 0.578 & 0.584  & 0.585      & 0.568                                                     & {\color[HTML]{FF0000} 0.632}  \\
                            & ClipIQA $\uparrow$ & 0.435                                                  & 0.440                         & 0.417    & 0.382                        &  0.562  & {\color[HTML]{FF0000} 0.578} & 0.528 & 0.483  & 0.502      & 0.509                                                     & {\color[HTML]{FF0000} 0.578}  \\
\multirow{-4}{*}{RealLQ250} & LIQE $\uparrow$    & 3.341                                                  & 3.280                         & 2.753    & 2.719                        & {\color[HTML]{2972F4} 4.005}  & 4.003                        & 3.904 & 3.605  & 3.688      & 3.639                                                     & {\color[HTML]{FF0000} 4.356} \\
\bottomrule
\end{tabular}
\caption{Quantitative comparison with state-of-the-art Real-ISR methods on four real-world benchmarks. Best and second best performance are highlighted in {\color[HTML]{FF0000} red} and {\color[HTML]{2972F4}blue}, respectively. Our DiT4SR achieves the best or comparable performance across four benchmarks. }
\vspace{-5mm}
\label{tab:results}
\end{table*}

Meanwhile,
the $3\times3$ depth-wise convolution layer also compensates for the limited local information-capturing ability of DiT.
As mentioned,
the joint attention is calculated at a global level,
thus lacking the local guidance of LR information.
Even if using one linear layer to inject LR guidance, 
it still remains challenging to repair fine structures such as letters,
as shown in \figref{fig:ffplus} (c).
The $3\times3$ depth-wise convolution helps capture more precise local information from the LR Stream and inject such LR guidance into the Noise Stream,
achieving better performance, especially for fine structures which are shown in \figref{fig:ffplus} (d).

\section{Experiments}
\subsection{Experimental Settings}
\myparaheadd{Datasets.}
Following SeeSR~\cite{wu2024seesr},
we adopt a combination of images from DIV2K~\cite{agustsson2017ntire}, DIV8K~\cite{gu2019div8k}, Flickr2K~\cite{timofte2017ntire}, and the first 10K face images from FFHQ~\cite{karras2019style} during training.
To fully exploit the potential of our method and scale up the training dataset,
another 1K self-captured high-resolution images are included.
The degradation pipeline of Real-ESRGAN~\cite{wang2021real} is utilized to synthesize LR-HR training pairs with the same parameter configuration as SeeSR.
Note that the resolutions are set to $128\times 128$ and $512 \times 512$ for LR and HR images, respectively.

Since our method specifically focuses on Real-ISR task,
we evaluate our model on four widely used real-world datasets,
including DrealSR~\cite{wei2020component}, RealSR~\cite{cai2019toward}, RealLR200~\cite{wu2024seesr}, and RealLQ250~\cite{ai2025dreamclear}.
All experiments are conducted with the scaling factor of $\times 4$.
DrealSR and RealSR respectively consist of 93 and 100 images.
Following SeeSR,
center-cropping is adopted for these two datasets,
and the resolution of LR images is set to $128 \times 128$.
RealLR200 is proposed in SeeSR, 
which comprises 200 images of significantly different resolutions.
RealLQ250 is established by DreamClear~\cite{ai2025dreamclear},
which consists of 200 images with a fixed resolution of $256 \times 256$.
Both RealLR200 and RealLQ250 lack corresponding GT images.

\myparaheadd{Metrics.}
As claimed in previous studies~\cite{jinjin2020pipal, blau2018perception, yu2024scaling},
full-reference metrics, like PSNR and SSIM~\cite{wang2004ssim},
are difficult to reflect the visual effects of restored results.
Thus,
most studies only use the perceptual measurement LPIPS~\cite{zhang2018unreasonable} for image fidelity.
We additionally include an evaluation of image fidelity in the user study.
We use MUSIQ~\cite{ke2021musiq}, MANIQA~\cite{yang2022maniqa}, ClipIQA~\cite{wang2023clipiqa}, and LIQE~\cite{zhang2023liqe} as non-reference metrics to measure image quality.

\myparaheadd{Implementation} is detailed in the supplementary material.

\begin{figure*}[t]
\begin{center}
\centerline{\includegraphics[width=.96\linewidth]{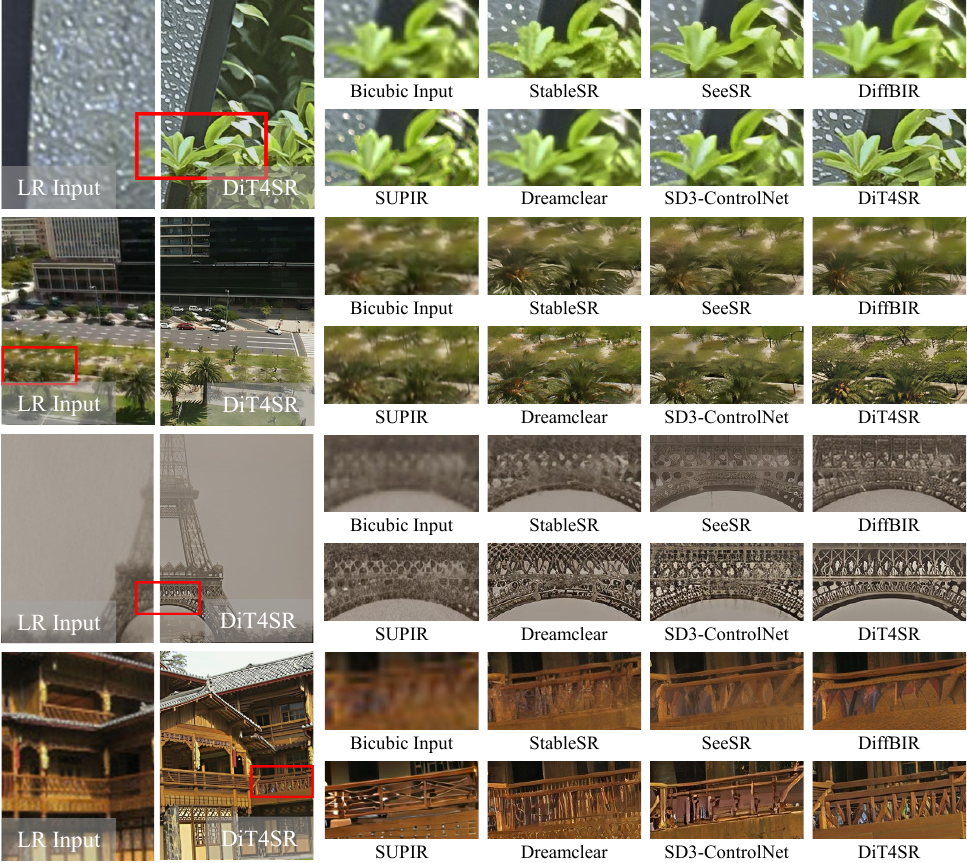}}
\caption{Qualitative comparisons with state-of-the-art Real-ISR methods on on RealSR (lst row), RealLR200 (2nd and 3rd rows), and RealLO250 (4th row). Our DiT4SR achieves the best performance in terms of image realism and detail generation while maintaining fidelity to the input LR image, especially preserving fine structures. More visual results can be found in the supplementary material. }
\label{fig:results}
\end{center}
\vspace{-11mm}
\end{figure*}

\subsection{Comparison with Other Methods}
We compare our method with state-of-the-art Real-ISR methods, including GAN-based methods (\ie Real-ESRGAN~\cite{wang2021real} and SwinIR~\cite{liang2021swinir}), diffusion-based methods with UNet architecture (\ie ResShift~\cite{yue2024resshift}, StableSR~\cite{wang2024exploiting}, SeeSR~\cite{wu2024seesr}, DiffBIR~\cite{lin2024diffbir}, OSEDiff~\cite{wu2024one}, and SUPIR~\cite{yu2024scaling}), and diffusion-based methods with DiT architecture (\ie DreamClear~\cite{ai2025dreamclear} and SD3-ControlNet). 
SD3-ControlNet is initialized with the SD3.5-medium parameters, employing the default configuration provided by Diffusers~\cite{von-platen-etal-2022-diffusers}
and trained under the same settings as our approach.

\myparaheadd{Quantitative Comparisons.}
The quantitative comparisons with state-of-the-art Real-ISR methods on four benchmarks are presented in \tabref{tab:results}.
For DrealSR and RealSR datasets, 
while SeeSR and DiffBIR demonstrate strong performance, 
our method achieves competitive results, closely matching or exceeding their performance. 
%
%
%
On RealLR200 and RealLQ250 datasets,
our method exhibits overwhelming performance, achieving top performance across all non-reference metrics.
These results highlight the capability of our method to produce high-quality restoration results.

\myparaheadd{Qualitative Comparisons.}
The qualitative comparisons against other methods are provided in \figref{fig:results}.
From the first two rows, 
our method is capable of generating results with better clarity and more abundant details over the compared methods, 
even when encountering severe blurring degradations.
This can be attributed to our method fully leveraging the outstanding generative capabilities of SD3.
Furthermore, as illustrated by the last two rows, 
our method exhibits a distinct advantage in processing fine structures, 
such as the architectural structure. 
Notably, even SD3-ControlNet, which is also built upon SD3, fails to handle these aspects as effectively.
This further highlights the superiority of our control mechanism over ControlNet, 
where more comprehensive information interaction enables the model to better leverage LR information, producing high-quality restoration outcomes with promising fidelity.

\myparaheadd{User Study.}
To further validate the restoration quality,
we invite 80 volunteers to conduct the user study.
We randomly select 60 LR images from these four datasets (DrealSR, RealSR, RealLR200, and RealLQ250),
and adopt the four latest methods (SeeSR, DiffBIR, SUPIR, and DreamClear) for comparison.
In the user study, 
participants were presented with three images for each evaluation: 
the original LR input image, the restoration result generated by our method, 
and the restoration result from a randomly selected other method. 
Participants were asked to answer two questions: (1) Which restoration result has higher image realism? (2) Which restoration result has better fidelity to the original image content?
%
%
The results reported in \tabref{tab:user_study} demonstrate that our method outperforms other approaches.

\begin{table}[tb]
\renewcommand{\arraystretch}{1.1}
\setlength{\tabcolsep}{6pt}
\centering
\footnotesize
\begin{tabular}{c|c|c|c|c}
\toprule
Ours vs.       & SeeSR     & DiffBIR     & SUPIR    & DreamClear    \\
\midrule
Realism       & 82.1\% & 83.6\%  & 81.7\%  & 72.7\%   \\
Fidelity      & 68.9\% & 79.5\%  & 75.4\%  & 64.5\%   \\
\bottomrule
\end{tabular}
\caption{Results of user study on real-world data. The numbers indicate the winning rate of our DiT4SR over the compared method in terms of image realism and fidelity.}
\label{tab:user_study}
\vspace{-2mm}
\end{table}

\begin{table}[t]
\footnotesize
\renewcommand{\arraystretch}{1.1}
\setlength{\tabcolsep}{4.5pt}
\centering
\begin{tabular}{c|ccc|cc}
\toprule
Model & \begin{tabular}[c]{@{}c@{}}LR\\ Integation\end{tabular} & \begin{tabular}[c]{@{}c@{}}LR\\ Residual\end{tabular} & \begin{tabular}[c]{@{}c@{}}LR\\ Injection\end{tabular} & MUSIQ $\uparrow$  & MANIQA $\uparrow$ \\ \midrule
FULL   &   \CheckmarkBold                                                      &   \CheckmarkBold                                                    &    \textbf{Conv}                                                     & 71.832 & 0.632  \\ \midrule
\rm{A}      &    \XSolidBrush                                                    &    \CheckmarkBold                                                   &     \textbf{Conv}                                                   & 66.963 & 0.574  \\
\rm{B}      &    \CheckmarkBold                                                    &     \XSolidBrush                                                  &     \textbf{Conv}                                                   & 70.887 & 0.614  \\
\rm{C}      &    \CheckmarkBold                                                     &     \CheckmarkBold                                                  &     \XSolidBrush                                                   & 71.202 & 0.610  \\
\rm{D}      &    \CheckmarkBold                                                     &      \CheckmarkBold                                                 &      \textbf{Linear}                                                  & 71.607 & 0.621 \\ \bottomrule
\end{tabular}
\caption{Ablation results on RealLQ250 for our DiT4SR. All
variants are trained using the same settings as the full model. }
\label{tab:ablation}
\vspace{-5mm}
\end{table}

\section{Ablation Study}
To further demonstrate the effectiveness of each component,
we conduct the ablation study on RealLQ250 with MUSIQ and MANIQA as evaluation metrics.
All variants are trained using the same settings as the full model for fair comparisons.
The detailed framework for each variant can be found in the supplementary material.

\begin{figure}[t]
\begin{center}
\centerline{\includegraphics[width=.98\columnwidth]{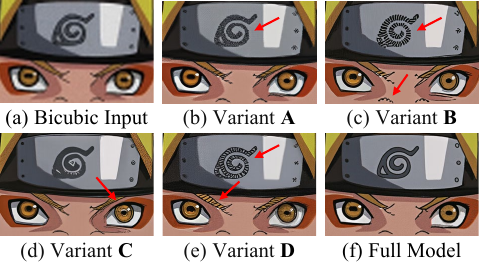}}
\caption{Visual comparison for the ablation study. Variant \textbf{A}, \textbf{B}, and \textbf{C} remove the LR Integation, LR
Residual, and LR Injection, respectively. Variant \textbf{D} replaces the convolution layer with the linear layer in LR Injection.}
\label{fig:ablation}
\end{center}
\vspace{-13mm}
\end{figure}

\myparaheadd{Effectiveness of LR Integration.}
To examine the effectiveness of LR Integration in Attention,
we exclude the LR Stream from the attention computation while keeping all other components unchanged.
%
From the results reported in \tabref{tab:ablation},
we can see that both MUSIQ and MANIQA present a significant decline,
which demonstrates that relying solely on LR injection between MLP layers is insufficient to generate high-quality results.
\figref{fig:ablation} (b) also indicates that without bidirectional information interaction between the LR Stream and the generated latent, degradations cannot be effectively removed.
This is because, without such information interaction, the LR guidance cannot adapt based on the evolving noisy latent,
which limits the ability of the model to adaptively refine the restoration.
Continuous interaction allows the LR information to progressively adjust and provide more accurate guidance, which is essential for effectively addressing complex degradations.

\myparaheadd{Effectiveness of LR Residual.}
Our intention of introducing LR residual is to maintain the consistency of LR guidance in deeper DiT blocks.
As shown in \tabref{tab:ablation},
without the LR residual, both MUSIQ and MANIQA exhibit a decline in performance.
From \figref{fig:ablation} (c),
we can see that the results contain several noticeable artifacts, which degrade the fidelity of the image content.
%
This issue primarily stems from the unstable evolution of the LR Stream,
which may encounter undesired disruptions.
%
By incorporating the LR Residual, 
our DiT4SR effectively stabilizes the LR Stream,
generating results with higher fidelity.

\myparaheadd{Effectiveness of LR Injection.}
First, we remove the LR Injection between MLP layers.
%
In \tabref{tab:ablation},
we can observe a slight decline in the evaluation metrics.
This indicates that relying on LR integration in attention alone can produce passable results.
However,
from~\figref{fig:ablation} (d), 
particularly in the eye region, 
we can observe noticeable content distortions.
This is mainly because attention operates globally, which is not sufficient for SR tasks that also rely on local information to accurately restore fine details.
We also replace the $3\times3$ depth-wise convolution with a linear layer.
Although \tabref{tab:ablation} shows similar performance between the linear layer and the convolution, 
\figref{fig:ablation} (e) indicates that artifacts and distortions have not been alleviated.
The 3 × 3 depth-wise convolution layer effectively compensates for the limited local information-capturing ability of DiT,
and significantly enhances the fidelity of our DiT4SR, 
which cannot be fully reflected by non-reference metric data.

\section{Conclusion}
In this paper, 
we present DiT4SR, 
one of the pioneering works to tame the large-scale DiT model for Real-ISR. 
Unlike ControlNet, which directly injects embeddings extracted from LR images, our method integrates LR embeddings within the original attention mechanism of DiT.
This enables bidirectional information flow between the LR latent and the generated latent.
Moreover, we introduce a cross-stream convolution layer that injects LR-guided information into the generated latent. 
This design not only enhances LR guidance but also compensates for DiT's weak local feature-capturing ability. 
Through these modifications,
DiT4SR achieves superior performance in Real-ISR, as demonstrated by extensive experiments. 
Our work highlights the potential of leveraging DiT for high-quality image restoration, paving the way for future research in this direction.